\documentclass[letterpaper, 10 pt, journal, twoside]{IEEEtran} 
\usepackage{amsmath,amsfonts}
\usepackage{algorithmic}
\usepackage{algorithm}
\usepackage{array}
\usepackage[caption=false,font=normalsize,labelfont=sf,textfont=sf]{subfig}
\usepackage{textcomp}
\usepackage{stfloats}
\usepackage{url}
\usepackage{verbatim}
\usepackage{graphicx}
\usepackage{cite}
\usepackage{xcolor}
\usepackage{graphicx}
\usepackage{svg}
\usepackage{amsthm}

\newtheorem{theorem}{Theorem}[section]
\newtheorem{definition}{Definition}[section]
\begin{document}

\title{Learning Cooperative Multi-Agent Policies with Partial Reward Decoupling}

\author{Benjamin Freed$^{*1}$, Aditya Kapoor$^{*1}$, Ian Abraham$^{2}$, Jeff Schneider$^{1}$, and Howie Choset$^{1}$
\thanks{Manuscript received: September 9, 2021; Revised: November 25, 2021; Accepted: November 29, 2021.} 
\thanks{This paper was recommended for publication by Editor Jens Kober upon evaluation of the Associate Editor and Reviewers’ comments.}  
\thanks{$^{1}$ These authors are with the Robotics Institute at Carnegie Mellon University, Pittsburgh, PA 15213, USA. {\tt\small \{bfreed,akapoor2,schneide,choset\}@cs.cmu.edu}.} 
\thanks{$^{2}$ Ian Abraham is with the Mechanical Engineering Department at Yale University, {\tt\small ian.abraham@yale.edu.}}}

\markboth{IEEE ROBOTICS AND AUTOMATION LETTERS. PREPRINT VERSION. ACCEPTED NOVEMBER, 2021}{Freed \MakeLowercase{\textit{et al.}}: Partial Reward Decoupling.}


\IEEEpubid{0000--0000/00\$00.00~\copyright~2021 IEEE}

\maketitle

\begin{abstract}
One of the preeminent obstacles to scaling multi-agent reinforcement learning to large numbers of agents  is assigning credit to individual agents' actions.  In this paper, we address this credit assignment problem with an approach that we call \textit{partial reward decoupling} (PRD), which attempts to decompose large cooperative multi-agent RL problems into decoupled subproblems involving subsets of agents, thereby simplifying credit assignment.  We empirically demonstrate that decomposing the RL problem using PRD in an actor-critic algorithm results in lower variance policy gradient estimates, which improves data efficiency, learning stability, and asymptotic performance across a wide array of multi-agent RL tasks, compared to various other actor-critic approaches.  Additionally, we relate our approach to counterfactual multi-agent policy gradient (COMA), a state-of-the-art MARL algorithm, and empirically show that our approach outperforms COMA by making better use of information in agents' reward streams, and by enabling recent advances in advantage estimation to be used. 
\end{abstract}

\begin{IEEEkeywords}
Reinforcement Learning, Multi-Robot Systems, Cooperating Robots
\end{IEEEkeywords}

\section{Introduction}
\IEEEPARstart{M}{any} difficult problems faced in robotics and AI involve large multi-agent systems, such as networks of autonomous vehicles, drone swarms, factory robot teams, or network routing.  Recently, multi-agent reinforcement learning (MARL) has shown promise on many of these problems, by enabling highly complex and cooperative strategies to be automatically \textit{learned}, as opposed to manually programmed.  For example, MARL algorithms have achieved super-human performance on complex multi-agent strategy games such as DOTA 2 \cite{openai5}, Starcraft II \cite{starcraft}, and capture the flag \cite{capture_the_flag}. 
    
    While MARL approaches have achieved impressive performance on these tasks, they require immense amounts of data and computing resources to train effectively, often requiring on the order of millions or billions of environmental interactions, necessitating costly GPU clusters.  The computational difficulty of large multi-agent RL (MARL) problems stems largely from the \textit{credit assignment problem}: identifying which agents' actions had a positive or negative impact on total group reward becomes increasingly difficult as the number of agents increases \cite{minsky,sutton_barto}.  
    
    \begin{figure}[!ht]
        \centering
        \includegraphics[width= \linewidth]{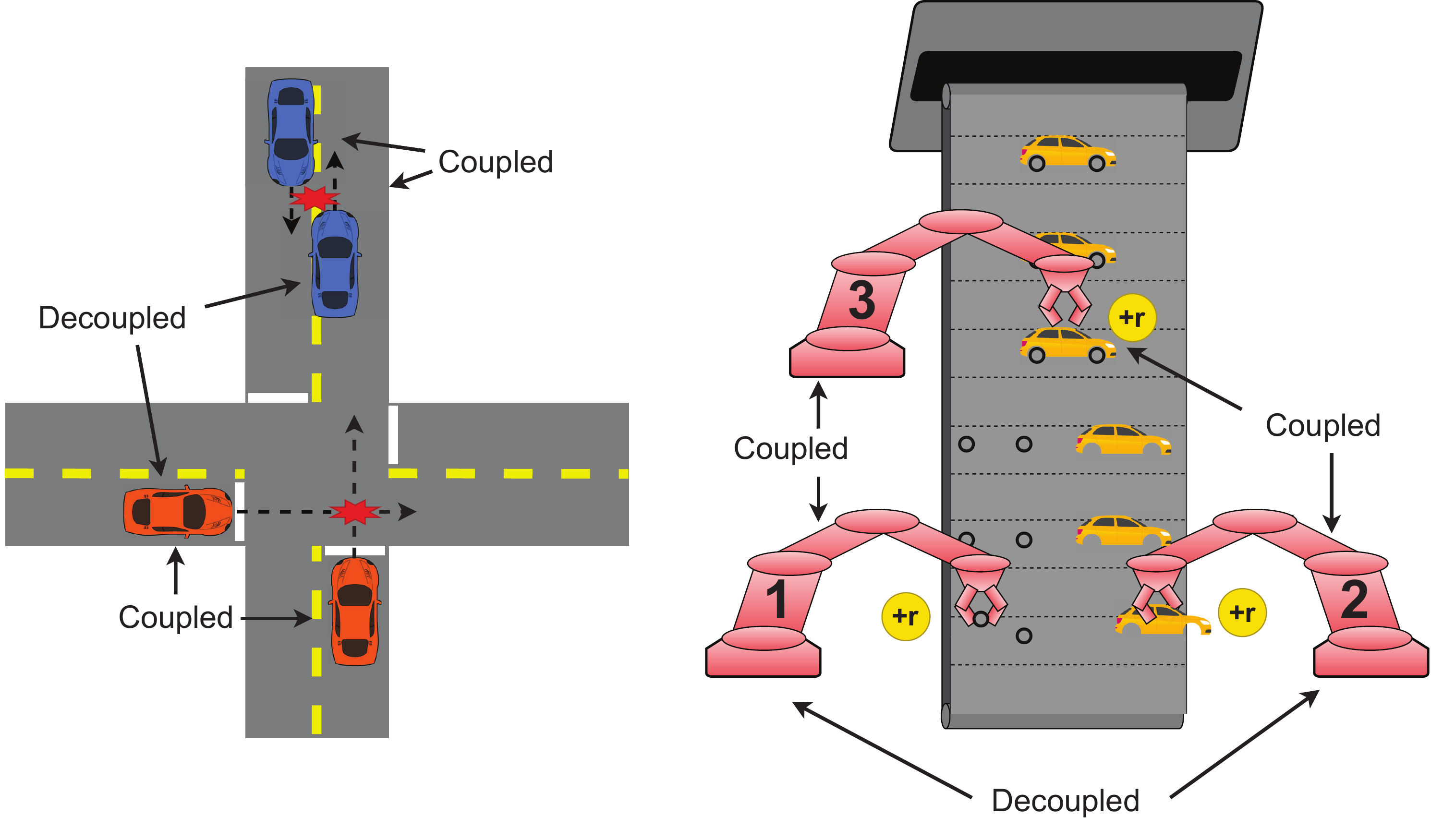}
        \caption{\textbf{Examples of problems in which PRD is applicable.}  \textit{Driving Scenario:} depicted in the left side, cars receive a negative reward for colliding.  The two orange and blue cars are on a collision course, meaning that each one's actions at the current timestep impact the future reward of the other, and they therefore cannot be decoupled using PRD. Note that neither orange car is at risk of colliding with either blue car, and therefore the orange and blue cars can be decoupled from one another using PRD.  \textit{Factory Scenario:} depicted in the right side, robot 1 is rewarded for placing wheels on the conveyor belt, robot 2 is rewarded for placing car bodies on the belt, and robot 3 is rewarded for successfully assembling the components.  Here, robots 1 and 2 may be decoupled using PRD as each one does not impact each other's rewards.  However, robot 1 and robot 2 cannot be decoupled from robot 3, because robot 3's ability to assemble cars and achieve reward is dependent on the actions of robots 1 and 2. }
        \label{fig:prd_diagram}
    \end{figure}
    In this paper, we address the difficulties associated with learning in large team sizes by streamlining credit assignment.  In model-free policy gradient-style algorithms (such as Actor Critic \cite{actor-critic}, Proximal Policy Optimization \cite{ppo}, Trust Region Policy Optimization \cite{trpo}, and Soft Actor-Critic \cite{sac}), we argue that the credit assignment problem manifests itself as a decrease in the signal-to-noise ratio in the learning update as the team size increases.  This viewpoint unifies two concurrent lines of work in model-free RL: credit assignment, and variance reduction.

   \IEEEpubidadjcol
    One way to combat the increased difficulty of credit assignment employs reward shaping \cite{reward_shaping}, wherein the reward function is meticulously hand-engineered by the algorithm designer to help guide the learning process.  For example, if it is known that an action taken by a particular agent has a deleterious impact on team performance, one may assign an explicit negative reward to that agent.  Many prominent MARL approaches incorporate some form of reward shaping.  However, reward shaping hardly offers a general solution to the credit assignment problem; for one, it may require a high degree of domain-specific knowledge, or tedious hyper-parameter tuning.  Additionally, a poorly designed reward function may limit the asymptotic performance of the team of agents, for example by making agents act selfishly or competitively when full cooperation is ideal.

    To address the credit assignment issue, we introduce a technique that we term \textit{partial reward decoupling} (PRD), which dynamically decomposes large groups of agents solving a cooperative task into smaller subgroups.  This decomposition is performed in such a way that if each agent learns how to optimally cooperate with agents within its subgroup, then the agents will also achieve optimal group-level cooperation.  Decomposing large multi-agent problems into smaller subproblems involving fewer agents reduces the number of agents among which credit must be distributed, thereby simplifying credit assignment.  Practically, within a policy gradient-style algorithm, we find that this streamlining of credit assignment results in lower variance policy-gradient estimates, therefore improving the data efficiency and stability of learning.

    PRD relies on the fact that, for many problems, it is the case that not every agent's actions are relevant to all other agents' rewards.  More precisely, if a particular agent's action only impacts the future rewards of a subset of other agents' individual rewards, then that agent need only explicitly attempt to cooperate with that subset of agents.  We show how this property can be used to improve the efficiency of the actor-critic algorithms \cite{actor-critic}, by filtering out the contributions of non-essential agents from policy gradient estimates. We use a value function with an attention mechanism to efficiently estimate which agents can be decoupled from one another.  We empirically show that our modified actor-critic algorithm (which we term \textit{partial reward-decoupled actor-critic}, PRD-AC) demonstrates more efficient and stable learning across a wide array of benchmarks, compared to conventional actor-critic, and a naive ``greedy" version of actor-critic, wherein each agent attempts to maximize its own individual reward.  The code used to perform all experiments is available \texttt{https://github.com/biorobotics/PRD}.

\section{Background}
\label{sec:background}

    \subsection{Multi-Agent Markov Decision Process}
    We consider multi-agent sequential decision-making problems that can be modeled as multi-agent Markov decision processes (MMDPs) containing \(M\) agents \cite{mmdp}.  An MMDP is defined by \( (\mathcal{S},\mathcal{A},\mathcal{P},\mathcal{R},\rho_0,\gamma ) \), where \(\mathcal{S}\) is the state space, \(\mathcal{A}\) is the joint action space, consisting of every possible combination of individual agents' actions, \(\mathcal{P}(s_{t+1}|s_t,a_t)\) specifies the state transition probability distribution, \(\mathcal{R}(r_t|s_t,a_t)\) specifies the reward distribution, \(\rho_0(s_0)\) denotes the initial state distribution, and \( \gamma \in (0,1]\) denotes a discount factor.  At each timestep, each agent \(i \in \{1,...,M\}\) selects an action independently according to its state-conditioned policy \( \pi_i(a_t^{(i)}|s^{(i)}_t;\theta_i) \), where \(s^{(i)}_t\) denotes the state information available to agent \(i\) (which may be the full state, as assumed in Sec. \ref{sec:results}), and \( \theta_i \) denotes the parameters for agent \(i\).  Each agent subsequently earns an individual reward \(r^{(i)}_t\) sampled from \(\mathcal{R}\) (that is, \(r^{(1)}_t,...,r^{(M)}_t \sim \mathcal{R}(\cdot|s_t,a_t)\)), and the environment undergoes a state transition, \(s_{t+1} \sim \mathcal{P}(\cdot|s_t,a_t)\).

    We focus on efficiently solving \textit{fully cooperative} tasks, wherein each agent attempts to maximize the sum of all agents' individual rewards.  More precisely, we wish to find the optimal agent policy parameters \(\theta^* = \{\theta_1^*,...,\theta_M^*\}\) that solve \( \theta^* = \underset{\theta}{\mathrm{argmax}}\ J(\theta) \), where
    \begin{equation}
        \label{eq:obj}
        J(\theta) = \mathbb{E} \Big[ \sum_{t=0}^T \sum_{j=1}^M \gamma^t r_t^{(j)} \Big],
    \end{equation}

    \noindent with \(T\) denoting the length of the episode.  By the policy gradient theorem \cite{pg}, the gradient of \(J\) with respect to the policy parameters \(\theta_i\) of a particular agent \(i\) is given by
    \vspace{-.1cm}
    \begin{multline}
        \label{eq:pg}
        \nabla_{\theta_i} J(\theta) = \mathbb{E} \Big[ \nabla_{\theta_i} \log \pi_i(a_t^{(i)}|s_t)  \\ \sum_{j=1}^M \left(Q_j^{\pi}(s_t,a_t)- b_{ij}(s_t,a_t^{- i}) \right) \Big],
    \end{multline}

    \noindent where \(Q_j^{\pi}(s_t,a_t)\) denotes the expected future reward for a particular agent \(j\), and \(b_{ij}(s_t,a_t^{- i})\) is commonly known as a \textit{baseline function}, which can be any any function that depends on the state, and the actions of all agents except \(i\) at time \(t\) \cite{action_dependent_baselines}.  \(b_{ij}\) cannot depend on the action of agent \(i\) because this would result in biased policy-gradient estimates.  Often a learned value function is used for the baseline.

    In the MMDP formulation we consider, agents earn \textit{individual rewards}; that is, each agent has its own unique stream of rewards.  Several past approaches to cooperative MARL have considered settings with \textit{global rewards}, in which the entire team earns one scalar reward value at each timestep.  We focus on the individual reward setting for two main reasons: 1) individual rewards are more \textit{general} than global rewards, and 2) individual rewards are relatively common in real-world settings.
    
    Individual rewards are more general than global rewards because any global reward function can be expressed as an individual reward function in which individual rewards across all agents are the same (\textit{i.e.}, \(r^{(i)}_t = r^{(j)}_t = r^{global}_t, \quad \forall i,j \in \{1,...,M\}\)). However, not all individual reward functions can be expressed as a global reward function, particularly in competitive or mixed cooperative-competitive problems such as those considered in \cite{maddpg}.  
    
    We also argue that individual rewards are relatively common in real-world settings.  For example, in vehicle routing, a natural reward function for a given agent is the rate at which it delivers goods to a customer.  Similarly, in a soccer game, individual rewards may be assigned to an agent that scores a goal.  Critically, we do not assume that a satisfactory cooperative solution may be achieved if each agent selfishly maximizes its own reward, which is why this paper proposes algorithms for maximizing \textit{total group reward}.  Note that credit assignment is still necessary even if individual rewards are assumed, because in general it remains unclear what the influence of a particular agent was on each agents' individual rewards.

\section{Partial Reward Decoupling}
\label{sec:prd}
    

    Many popular RL algorithms, such as SAC \cite{sac}, PPO \cite{ppo}, TRPO \cite{trpo}, D4PG \cite{d4pg}, MADDPG \cite{maddpg}, and COMA \cite{coma}  work by estimating the gradient specified in eq. \eqref{eq:pg}.  However, in a MARL context, these gradient estimates are often very noisy due to the fact that they are essentially a sum over contributions from (potentially many) agents, each of which contributes noise to the policy gradient estimate.  Luckily, it is often the case that large-scale cooperative multi-agent problems can be thought of as a collection of decoupled subproblems, involving only a fraction of the entire group.  For example, in the robot assembly problem depicted in the right of Figure \ref{fig:prd_diagram}, imagine that robots 1 and 2 are both rewarded for placing items on the belt, and do not influence the quality with which the other can perform its job, when functioning properly.  Therefore, robots 1 and 2 may be \textit{decoupled} (that is, treated as belonging to separate subteams), and do not need to explicitly cooperate.  However, both robots 1 and 2 influence the quality with which robot 3 completes a downstream process, and therefore neither robots 1 or 2 can be decoupled from robot 3.  If robot 1 were to malfunction and interfere with robot 2's ability to perform its function by blocking it, this would also introduce coupling between robots 1 and 2.
    
    While in the robot assembly example the possible decouplings are fairly obvious, for many problems the decoupling is non-obvious, and may vary with time. For example, in the driving scenario depicted on the left of the figure, for a portion of the time it may be possible to decouple an agent from all other agents.  For example, if we are confident that two vehicles do not influence each other at a particular time, or any time in the future, the vehicles may be decoupled and treated as parts of two separate RL problems.  However, when a vehicle enters a busy intersection, or is at risk of colliding with another vehicle, it should be considered coupled to other vehicles, at least momentarily.  
    
    Our proposed approach, PRD, decomposes multi-agent problems based on which agents' actions impact the future reward of other agents.  If a particular agent \(i\) does not impact the expected future reward of another agent \(j\), then agent \(i\) can be \textit{decoupled} from agent \(j\).  In the context of a policy gradient-style algorithm, this decoupling means that  the contribution of agent \(j\)'s rewards can be removed from the gradient estimate of agent \(i\), without introducing bias.  We claim PRD addresses the credit assignment problem, because credit for agent i's rewards is assigned only to those agents that had an influence on agent i's rewards.  PRD can also be seen as a variance reduction technique, because in a policy gradient-style algorithm, we expect this decoupling to result in lower variance policy gradient estimates, as each policy gradient estimate would consider only the contributions of a smaller subset of agents, thus reducing the amount of noise that is injected.

    \begin{definition}
    The subset of agents that are impacted by the action of agent \(i\) at time \(t\) is defined as the \textit{relevant set} of agent \(i\), \(R^{\pi}_i(s_t,a_t)\), which is defined implicity such that 
    
    \vspace{-.3cm}
    \begin{multline}
        \label{eq:relevant_set_def}
        \mathbb{E}\big[\sum_{\tau=t}^T r_\tau^{(j)} | s_t,a_t \big] =  \\
        \mathbb{E}\big[\sum_{\tau=t}^T r_\tau^{(j)} | s_t,\{a^{(i)}_t | i \in \{1,...,M\}, j \in R_i^{\pi}(s_t,a_t) \} \big].
    \end{multline}
    That is, the expected future reward of a particular agent \(j\), given the current state of the environment, only depends on the actions of agents that include \(j\) in their relevant set.  Note that \(R_i^{\pi}\) changes with time.
    \end{definition}

    We include the joint policy \(\pi\) in the superscript because the relevant set depends on the joint policy of all the agents.
    \begin{theorem}
    \label{theorem:prd}
    Using the definition of the relevant set presented in \eqref{eq:relevant_set_def}, the gradient update of agent \(i\) can be simplified by removing the contributions of agents not in the relevant set of agent \(i\).  That is, the policy gradient of agent \(i\) can be simplified to (omitting the arguments of the Q functions, baselines, and the relevant set for brevity)
    \vspace{-.1cm}
    \begin{equation}
        \label{eq:simplified_pg}
        \nabla_{\theta_i} J(\theta) = \mathbb{E}\Big[ \nabla_{\theta_i} \log \pi_i(a_t^{(i)}|s_t) \sum_{j \in R^{\pi}_i} \big( Q_j^{\pi} - b_{ij} \big) \Big].
    \end{equation}    
    \end{theorem}
    \begin{proof}
    If agent \(j\) is outside of the relevant set of agent \(i\), then \(Q^{\pi}_j\) does not depend on \(a_t^{(i)}\), and can therefore be subtracted out as an action-independent baseline without biasing the policy gradient.  That is, \(b_{ij}(s_t,a_t^{- i}) = Q^{\pi}_j\), given that \(j \notin R^{\pi}_i(s_t,a_t)\).  Therefore, terms corresponding to \(j \notin R^{\pi}_i(s_t,a_t)\) do not contribute to summation in \eqref{eq:pg}, and may therefore be removed.
    \end{proof}
    
    Omitting the contributions of irrelevant agents from the learning update of agent \(i\) can be seen as decomposing a (potentially very large) cooperative multi-agent problem involving \(M\) agents into a set of smaller ones, involving only \(|R_i(s_t,a_t)|\) agents, at each timestep \(t\).  However, unlike a static decomposition that may attempt to assign agents to static subgroups, partial reward decoupling assigns agents to subgroups that are \textit{overlapping} (\textit{i.e.}, an agent may belong to multiple subgroups simultaneously) and \textit{dynamic} (\textit{i.e.}, relevant set assignments change from timestep to timestep).  In the following section we explain how we use learned value functions with attention mechanisms to efficiently infer the relevant set for each agent.

    \subsection{Estimating the Relevant Set}
    \label{subsec:relevant_set}
    
    In practice, because the dynamics of the environment and reward function are unknown to agents, the relevant set must be estimated empirically through interactions with the environment.  To estimate the relevant set of each agent, we utilize the common technique of value function estimation \cite{pg,maddpg,actor-critic,actor-attention-critic}.  During training, we maintain a learned value function \(V^{\pi}_{ij}\) that attempts to estimate the expected individual future reward for each agent \(j\), given the current state and the actions of all agents \textit{except} \(i\), which we explain further in Sec. \ref{subsec:ac_arch}.  We acknowledge that this definition of a value function differs from past works where value was conditioned only on the state, we find this extension useful because it enables us to use \(V^{\pi}_{ij}\) both as a baseline \(b_{ij}\), and for relevant set estimation.  While it may appear that estimating \(V^{\pi}_{ij} \quad \forall i,j\) requires \(M^2\) different learned value functions (one for every combination of \(i\) and \(j\)), in practice we find that we can efficiently estimate \(V^{\pi}_{ij}\) for any \((i,j)\) with \textit{one} set of value function parameters, using the architecture described in Sec. \ref{subsec:ac_arch} and by sharing weights across agents.  
    
    Recall that, if the expected future reward of agent \(j\) does not depend on the action of agent \(i\) at a particular time \(t\), then \(j\) is not in agent \(i\)'s relevant set at time \(t\).  This property justifies the use of the value function to estimate the relevant set; if our value estimate for agent \(j\) does not depend on \(a_t^{(i)}\), we infer \(j \notin R^{\pi}_i(s_t,a_t)\).  We use a value function with an attention mechanism over agent actions, which allows dependence on particular agents' actions to be ``shut off'' as a function of the current state of the environment, by setting the attention weight for the corresponding action to 0.

    In the following section, we use this observation to modify the popular actor-critic algorithm \cite{actor-critic} to help solve the credit assignment problem, by selectively removing the contributions of agents not within the relevant set from estimated policy gradients.


\section{Partial Reward Decoupled Actor-Critic}
\label{sec:prd-ac}

To create an efficient MARL algorithm using PRD, we combine PRD with the actor-critic algorithm \cite{actor-critic}, which we refer to as PRD Actor-Critic (PRD-AC). In PRD-AC, value function estimation is used to determine which advantage terms may be eliminated from the learning updates.  Here, PRD can be seen as a variance reduction technique, because it removes unnecessary advantage terms that inject noise into the gradients.  
\subsection{Actor and Critic Architectures}
\label{subsec:ac_arch}

Each agent maintains identical copies of an actor network and a critic network.  The actor network is a multi-layer perceptron (MLP), which takes the local observation of the current agent, concatenated with all other agents' local observations.  To support decentralized execution, local observations along with messages from other agents may be provided as input to the  actor network instead \cite{sparse_comms,magnet}.  The critic network is a graph neural network (GNN) with a graph attention mechanism \cite{gat} that outputs \(V^{\pi}_{ij}\).  We do not condition \(V_{ij}^{\pi}\) on the action of agent \(i\), so that in addition to estimating the relevant set, \(V_{ij}^{\pi}\) can be used as a baseline (\(b_{ij}\) form eq. \eqref{eq:pg}) \cite{actor-critic,action_dependent_baselines}.  Note that, in general, \(V_{ij}^{\pi} \neq V_{kj}^{\pi}\) for some \(k \neq i\), because \(V_{ij}^{\pi} \) and \(V_{kj}^{\pi}\) condition on the actions of different sets of agents.

The graph attention mechanism used in the critic network is based on scaled dot product attention \cite{attention_is_all_you_need}.  For each agent (corresponding to a node in the GNN), at each timestep, state components pertaining to individual agents (\textit{i.e.}, position, velocity, and goal position) are pre-processed using a MLP, and then used to compute keys and queries.  Attention values for each agent \(k\) are computed using the pre-processed state and action of agent \(k\), with the exception that \(a_t^{(i)}\) is replaced with \(\pi^{(i)}(s_t)\), to avoid conditioning agent \(i\)'s baseline \(V_{ij}^{\pi}\) on its own action.  Keys, queries, and values are subsequently passed through a single-head self-attention (SSA) mechanism, which combines attention values from each agent according to a weighted sum \cite{attention_is_all_you_need}.  The output of the SSA is passed through a linear layer to compute \(V^{\pi}\), an \(M\)-by-\(M\) matrix of value estimates \(V_{ij}^{\pi}\) for each \(i,j \in \{1,...,M\}\).  The SSA also outputs the weight matrix \(W\), where \(W[k,j]\) (which we refer to as \(w_{kj}(s_t)\)) indicates how much agent \(j\) attends to the observation-action of agent \(k\) while computing \(V_{ij}^{\pi}\), for each possible value of \(i\) (that is, \(V_{ij}^{\pi}\) places the same attention weight on the observation-action for agent \(k\) when estimating the expected future rewards for agent \(j\), regardless of \(i\).  While theorem \ref{theorem:prd} motivates using attention weights over actions only, we found that attending over observations and actions stabilized training.  

\subsection{Actor Update}
\label{sec:actor_update}

The policy update for agent \(i\) in PRD-AC is similar to that of a standard actor-critic algorithm, with the primary exception being that advantage terms outside of agent \(i\)'s relevant set are removed.  Note that, because \(w_{ij}\) varies with time, so does the estimated relevant set, enabling decoupling to be dynamic with time.  After each episode, we compute one gradient-based policy update.  The gradient estimate for the policy parameters of  agent \(i\) is given by
\begin{equation}
    g_i = \sum_{t=0}^T\nabla_{ \theta_i} \log \pi_i(a^{(i)}_t|s_t) \sum_{j: w_{ij}(s_t) > \epsilon} \hat{A}_{ij}(s_t,a_t), 
\end{equation}

\noindent where \(\hat{A}_{ij}(s_t,a_t)\) is the advantage estimate computed using the actual rewards of agent \(j\), and the critic-estimated expected rewards of agent \(j\), not conditioned on the actions of agent \(i\).  There are many ways this advantage may be estimated, but we choose to use generalized advantage estimation (GAE) \cite{gae}, with \(\lambda=0.98\).  Here, \(\epsilon\) is the threshold above which we consider \(w_{ij}\) to be relevant.  In practice, we initialize \(\epsilon\) to 0 (making the algorithm initially equivalent to standard actor-critic), and linearly increase it over the first 15K episodes to a maximum value of 0.01, which we found empirically to stabilizes training, by allowing the critic time to learn how to correctly assign attention weights.

\subsection{Critic Update}

The output of the critic network is regressed against the TD(\(\lambda\)) target value estimate, with \(\lambda=0.8\) \cite{sutton_barto}.  After each episode of environmental interaction, we apply one gradient update to the parameters of the critic network.  The gradients used to update the critic loss are obtained by differentiating the Huber loss \cite{huber_loss} between the critic network outputs and TD(\(\lambda\)) value estimates \cite{sutton_barto} for the last episode, which is used to compute the critic learning update.  Huber loss is similar to the more commonly-used MSE, but is more robust to outliers, which we found stabilized training.  

    \subsection{Relation to Counterfactual Multi-Agent Policy Gradient}
    Counterfactual multi-agent policy gradient (COMA) is another actor-critic algorithm that addresses the credit assignment problem, and to the best of our knowledge is the existing algorithm most closely related to PRD-AC.  PRD-AC retains the benefits of COMA, and extends them in two key ways: 1) when individual rewards are available, PRD-AC can take advantage of the additional information contained in the individual reward streams to further streamline the gradient updates, and 2) PRD enables a wide range of advantage estimation techniques to be used (for example, GAE, which we use in our experiments).  In practice, we find that both of these advantages yield substantial improvements in learning speed and asymptotic performance (see sec. \ref{subsec:coma_to_prd}).
    
    Adapting notation to fit the rest of the paper, the COMA gradient estimator can be expressed as 
    \begin{equation}
        \label{eq:coma_grad}
        g_i^\text{COMA} = \nabla_{\theta_i} \log \pi_i(a_t^{(i)}|s_t^{(i)}) A^{(\text{COMA},i)}(s_t,a_t),
    \end{equation}
    \noindent where the COMA advantage estimate is given by
    \begin{equation}
    \label{eq:coma_adv}
    A^{(\text{COMA},i)}(s_t,a_t)=Q^{\pi}(s_t,a_t) - \mathbb{E}_{a_t^{(i)}} [Q^{\pi}(s_t,a_t)].  
    \end{equation}
    
    COMA has a similar benefit to PRD-AC in that, if the action of agent \(i\) does not impact the expected future group reward, its influence from the gradient estimate is eliminated.  This is captured implicitly in COMA by virtue of the fact that the COMA advantage estimate \(A^{(COMA,i)}\) is \(0\) when agent \(i\)'s action does not impact expected future group reward (\textit{i.e.}, when \(Q^{\pi}(s_t,a_t) = \mathbb{E}_{a_t^{(i)}} [Q^{\pi}(s_t,a_t)\).  This effect is captured in PRD-AC because if agent \(i\)'s action does not impact group rewards (that is, if  \(Q_j^{\pi}(s_t,a_t) = \mathbb{E}_{a_t^{(i)}} [Q_j^{\pi}(s_t,a_t)], \quad \forall j\), then agent \(i\)'s relevant set is empty, and agent \(i\)'s advantage terms are omitted from the gradient estimate.  However, PRD-AC extends COMA by making use of additional information in individual agents' reward streams, if they are available.  While COMA can only eliminate advantage terms when the corresponding agents' actions have no impact on \textit{total expected group reward},
    PRD-AC can eliminate advantage terms on an individual-agent basis; for example, if agent \(i\)'s action does not impact the expected future reward of agent \(j\), then \(\hat{A}_{ij}\) may be eliminated.  The ability to remove advantages on an individual-agent basis is beneficial because often it may be the case that a particular agent's action impacts only a subset of other agents' rewards, allowing us to streamline that agent's learning update, despite the fact that agent \(i\)'s action may still impacts future group reward, for instance by impacting other agents besides \(j\).
    
    The second way PRD-AC extends COMA is by enabling a wider range of advantage estimation techniques to be used.  While COMA assumes advantage estimates of the form shown in eq. \eqref{eq:coma_adv}, PRD-AC places no restriction on the advantage estimation technique used, and we find that other advantage estimation techniques (in particular, GAE) offer improved performance. 
    

\section{Experimental Results}
\label{sec:results}
We experimentally evaluate the performance of PRD-AC on a range of benchmarks designed to require various levels and types of cooperation (e.g., cooperation based on teams, or based on physical proximity).  We compare average episode reward vs. elapsed training episodes PRD-AC to COMA, as well as two other baseline actor-critic algorithms, one that encourages full cooperation (Shared-AC), and one that encourages agents to act selfishly (Greedy-AC).  To validate our claim that PRD-AC reduces policy gradient estimator variance, we compare the variance of gradient estimates of PRD-AC to an actor-critic algorithm in which no decoupling occurs (Shared-AC).  Finally, to better understand the behavior of PRD-AC, we perform a series of ablation experiments that interpolate between PRD-AC and COMA.


\subsection{Comparison to Greedy and Shared-Reward Actor-Critic}
\label{subsec:greedy_shared}
To evaluate the effect of PRD on learning efficiency and asymptotic performance, we compare our PRD-AC algorithm (see Section \ref{sec:prd-ac}) to the following baseline approaches on a set of multi-agent RL tasks:

\begin{itemize}
    \item \textbf{Greedy Actor-Critic (Greedy-AC)}: each agent attempts to maximize its own individual reward.
    \item \textbf{Shared-Reward Actor-Critic (Shared-AC)}: each agent attempts to maximize the sum of group rewards.
    \item \textbf{COMA.}
\end{itemize}

We considered greedy-AC as a baseline because it represents a naive form of reward shaping, that may work well for many problems in which little to no cooperation is required.  Greedy-AC simplifies the learning problem, because each agent need not consider its impact on the rewards of other agents.  However, it does not encourage cooperation among agents.  We additionally considered shared-AC as a baseline because a shared reward function in which every agent optimizes its policy for total group reward should discover fully cooperative policies that are optimal on a group level.  The primary drawback to shared-reward AC is that it has a higher variance learning signal, because it must consider the rewards from all agents. We hypothesize that PRD-AC will combine the benefits of both greedy and shared-reward actor-critic by allowing each agent to selectively consider only the rewards of other agents that are impacted by its behavior in its learning updates.  Intuitively, for each agent \(i\), PRD-AC attempts to find the smallest set of other agents that agent \(i\) must cooperate with (\textit{i.e.,} its relevant set) so as to to yield group-level solutions; in essence, PRD-AC acts as greedily as possible while still being fully cooperative. 

We evaluate PRD-AC, greedy-AC, and shared-AC in three distinct task settings, which we refer to as \textit{paired-agent goal finding, collision avoidance,} and \textit{social dilemma}.  Each task requires agents to navigate from randomly-selected start locations to randomly-selected goal locations.  In each task, agents modeled as point masses in a 2-dimensional planar environment.  The chosen evaluation tasks require varying degrees of cooperation; additionally, each task involves unique strategies for ideal relevant set selection.  For each experiment, we run 5 distinct trials of each algorithm (unless otherwise indicated) with distinct random seeds.  We plot the mean episode reward across all trials, as well as a shaded region indicating 1 standard deviation above and below the mean.

\textbf{Paired-Agent Goal-Finding:} agents are randomly paired with exactly 1 other partner agent.  Agents must navigate from their initial location to their goal location, but are not rewarded themselves for reaching their goal.  Instead, each agent is rewarded based on how much closer their partner gets to its goal.  This task is intended to be a somewhat contrived toy problem,  designed to present a situation in which a greedy approach fails catastrophically because each agent's individual reward is independent of its actions.  Therefore, the task cannot be effectively solved by having every agent attempt to maximize its own individual reward. 

\begin{figure*}[t]
    \vspace{.2cm}
    \centering
    \includegraphics[width=\textwidth]{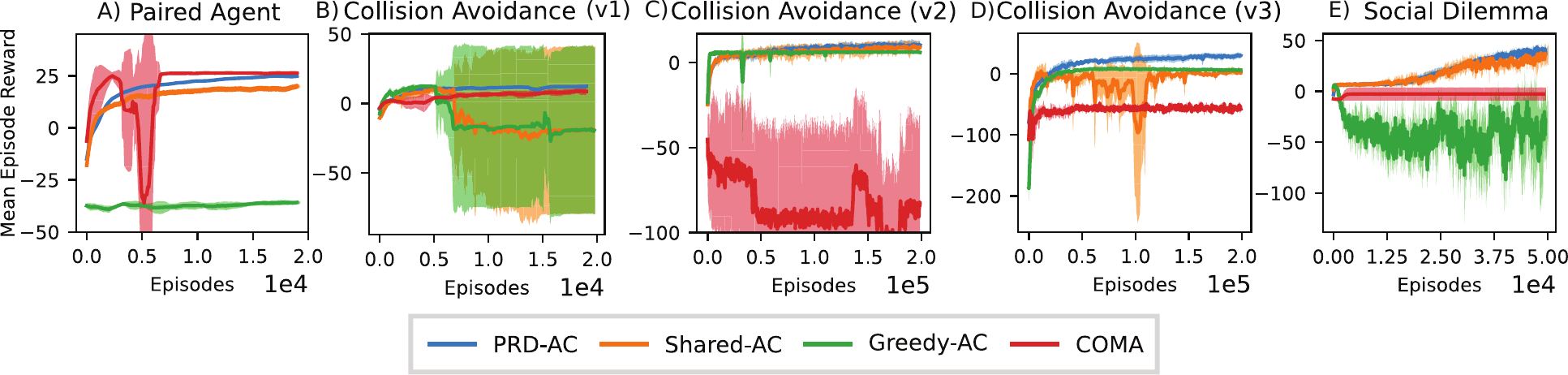}
    \caption{\textbf{Mean episode reward vs. episodes for paired agent, collision avoidance, and social dilemma tasks.}  PRD-AC reliably solved all of the tasks, and tended to outperform the other algorithms we tested.  COMA tended to suffer from instability despite extensive hyperparameter tuning, and underperformed PRD-AC in all but the simplest (paired agent) environment.  Neither Shared-AC nor Greedy-AC reached the same level of asymptotic performance as PRD-AC on most tasks, with Greedy-AC failing to make progress on tasks that required high degrees of cooperation, such as paired agent and social dilemma.}
    \label{fig:reward_plots}
\end{figure*}

We compare PRD-AC, greedy-AC, and shared-AC in a task with 30 agents.  As predicted, greedy-AC fails to make significant progress towards solving the task (Fig. \ref{fig:reward_plots}).  PRD-AC outperforms the shared reward strategy, achieving a higher reward more rapidly.  COMA learned more rapidly than PRD-AC, but suffered from instability.


\textbf{Collision Avoidance:} Agents are spawned on the perimeter of a square planar environment, and must navigate to a goal location diametrically opposite to them, while avoiding collisions with other agents.  Agents' rewards are composed of a component that is positive when agents move closer to their goals, and a negative collision penalty.  We consider three distinct methods for assigning collision penalties:

\begin{enumerate}
    \item Collision penalties on colliding agents (v1): any two agents that collide are both penalized (M=8).
    \item Collision penalties on non-colliding agents (v2): all agents \textit{except} the colliding agents are penalized (M=8). 
    \item Collision penalties on non-colliding team-members (v3): Agents are divided into 3 teams of 8 agents. When two agents of the same team collide, all other non-colliding agents of the same team are penalized. 
\end{enumerate}

Assigning collision penalties to colliding agents can be seen as a form of reward shaping, wherein the offending agent is directly penalized for impeding the progress of other agents.  We expect this task to favor greedy-AC, since individual agents' rewards already encourage some amount of cooperation.  On the other hand, assigning collision penalties to non-colliding agents (as is done in v2 and v3) simulates \textit{externalities}, a concept from economics wherein an agent's actions have a negative impact on other agents without negatively affecting the offending agent.  In this case, we expect greedy-AC to fail, because agents' individual rewards do not encourage them to avoid collisions, which we hypothesize will result in selfish behavior.  We expect shared-AC outperform greedy-AC, since agents attempt to maximize group reward rather than individual reward.  However, we expect PRD-AC to outperform both greedy-AC and shared-AC because it should learn cooperative policies while benefiting from lower-variance gradient updates compared to shared-reward actor-critic.

In v1, while greedy initially learns most rapidly out of the four approaches, and shared reward appears to match the performance of PRD, both greedy and shared suffer from learning instability and experience a reward crash just prior to convergence.  In the case of shared-AC, we hypothesize that this crash is due to high-variance gradient estimates.  COMA learns more slowly than the other 3 approaches.  In v2, we find that greedy-AC learns a suboptimal policy as predicted, and that PRD-AC slightly outperforms shared-AC.  COMA suffers from instability and fails.  In v3, PRD-AC outperforms shared-AC and greedy-AC by a much larger margin than in v2; we hypothesize that this is due to the fact that in v3, each agent's actions impact a smaller number of other agents, allowing decoupling to be more effective.  COMA converges to a sub-optimal solution in this experiment.

\textbf{Social Dilemma:} 2 teams of agents each are spawned in random locations on one side of a square and 2 goal regions (one for each team) are placed on the opposite side of the square. Agents receive a positive reward for moving closer to \textit{either} of the goals, and an additional positive reward for being within a goal region.  Agents are incentivized to move as quickly as possible to a goal location, because rewards earned further in the future are discounted more heavily. However, if an agent enters a goal region that belongs to the other team, every agent belonging to the other team receives a negative penalty. The strategy which maximizes total group reward is for each agent to travel to the goal associated with its team; however, the strategy that maximizes individual reward is for each agent to go to the goal to which it is closest initially, which will be the opposite team's goal roughly 50\% of the time, thereby imposing a penalty on the other team and lowering group reward.  We therefore expect greedy-AC to learn a suboptimal strategy.

As predicted, greedy learns a severely suboptimal strategy. PRD achieves a higher maximal reward than shared.  COMA converges to a suboptimal strategy.

\begin{figure}[t]
    \vspace{.2cm}
    \centering
    \includegraphics[width=.8 \linewidth]{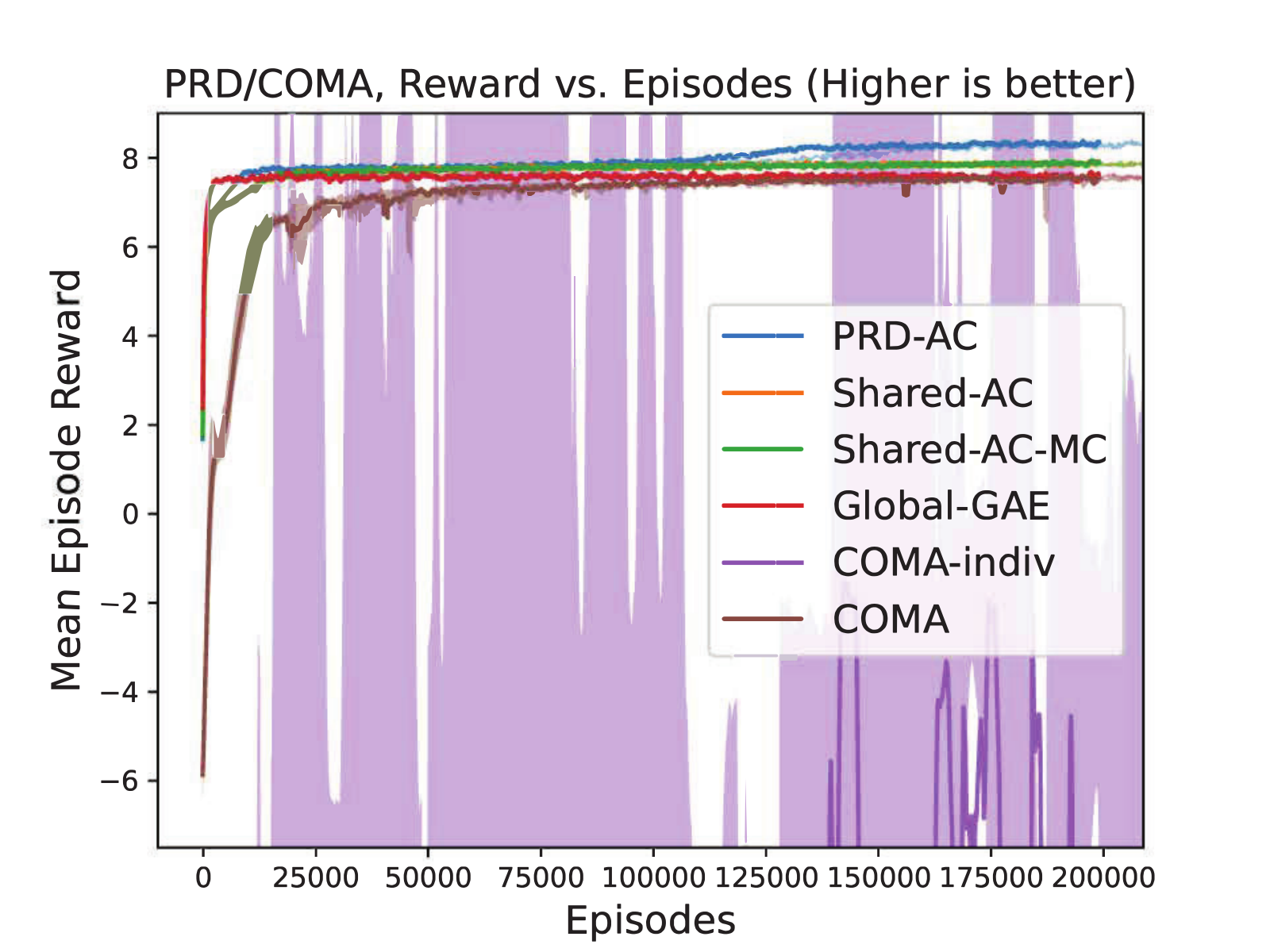}
    \vspace{-0.3cm}
    \caption{Results for COMA/PRD comparison.  PRD-AC outperforms all other algorithms.  COMA is outperformed by global-GAE, indicating that COMA's advantage estimation is suboptimal.}
    \label{fig:coma_to_prd}
    \vspace{-.5cm}
\end{figure}

\vspace{-.15cm}
\subsection{Policy Gradient Estimator Variance}
\label{subsec:gradvar}

To better understand the reasons for the performance improvement of PRD-AC over shared-AC, we plot the variance of the policy gradient estimates for both PRD-AC and shared-AC at various points in the training process for the paired agent task with M=30.  Because PRD-AC is able to eliminate the stochastic advantage estimates of agents outside of the relevant set from its policy gradient estimate, and these advantage estimates contribute noise to the gradient estimates, we hypothesize that the gradients estimated by PRD-AC will have a significantly lower variance than those estimated by shared-AC.

For maximal comparability between the two different gradient estimation techniques, we use the identical actor and critic networks (\textit{i.e.,} with the same parameters) to compute the gradient variance estimates for both PRD-AC and shared-AC.  That way, any differences in variance can only be explained by differences in gradient estimation technique, rather than differences in actor or critic network parameters, differences in policy stochasticity, or any other confounding variables that may influence the variance of gradient estimates.  When computing the gradient variance for both PRD-AC and shared-AC, we use actor and critic network parameters that were saved every 1000 episodes during one run of shared-AC.  We estimate variances empirically by taking the variance across 100 independent gradient estimates, each computed using data from an independent episode.

\begin{figure}[t]
    \centering
    \includegraphics[width=.8 \linewidth]{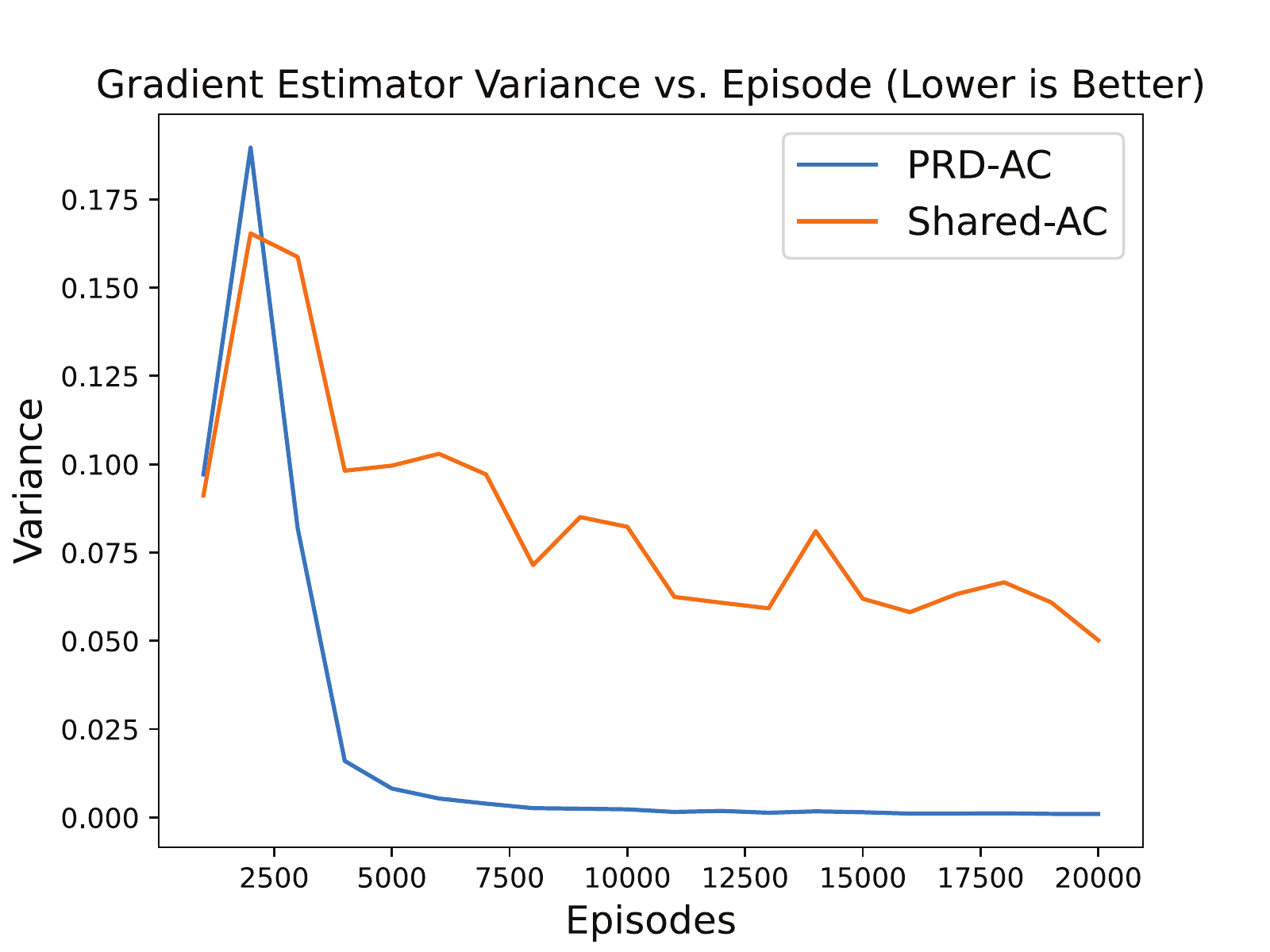}
    \caption{Comparison of gradient estimator variance between PRD-AC and shared-AC.  PRD-AC has significantly lower gradient estimator variance than shared-AC.}
    \label{fig:gradvar}
    \vspace{-.5cm}
\end{figure}

We observe that our hypothesis that PRD-AC provides lower variance gradient estimates is correct (Fig. \ref{fig:gradvar}), especially in later episodes, where the magnitude of the variance differs by over an order of magnitude.  These observations give a strong indication that the reduction in variance enabled by PRD is responsible for the improvements in the learning efficiency and stability of PRD-AC over shared-AC.

\subsection{Comparison to COMA}
\label{subsec:coma_to_prd}

To better understand the advantages of PRD-AC over COMA, we run both algorithms on the same task (paired agent task with M=8), and additionally perform a series of ablation experiments that combine different aspects of the two algorithms.  The experiments are labeled as follows:

\begin{itemize}
    \item Shared-AC-GAE: "default" implementation of shared-AC used to generate the results in sec. \ref{subsec:greedy_shared}.  Shared-AC-GAE is equivalent to PRD-AC, except all advantage terms are retained in the gradient update.
    \item Shared-AC-MC: Equivalent to shared-AC-GAE, except Monte-Carlo (MC) advantage estimates are used; that is, advantages are estimated by subtracting a learned value-function baseline (identical to the one used in PRD-AC and Shared-AC-GAE) from a Monte-Carlo estimate of expected future returns (that is, simply the sum of future returns for one episode). 
     \item Global-GAE: Equivalent to COMA (uses global rewards) but estimates advantages using GAE instead of COMA's advantage estimation technique. Note that we do not consider this algorithm to be a variant of COMA, because COMA is essentially defined by a particular method of advantage estimation.  An additional difference between Global-GAE and COMA is that COMA fits a Q-function while Global-GAE fits a value function.
    \item COMA-indiv: Equivalent to COMA, except Q functions are decomposed on an individual-agent basis.  COMA-indiv is also equivalent to shared-AC-GAE and shared-AC-MC, except for the fact that COMA-style advantage estimates are used, as opposed to GAE or MC advantage estimates.  This experiment represents an adaptation of COMA to a setting with individual rewards.
    \item COMA \cite{coma}.
\end{itemize}

We find that PRD-AC significantly outperforms COMA, with PRD-AC achieving higher reward than COMA during the entirety of training (Fig. \ref{fig:coma_to_prd}).  We additionally find that shared-AC-GAE, shared-AC-MC, and global-GAE all outperform COMA, despite not utilizing any sophisticated credit-assignment schemes, suggesting that the advantage estimation scheme used by COMA could be improved.  Interestingly, global-GAE outperforms COMA, because global-GAE is equivalent to COMA other than the fact that it uses GAE. We hypothesize that COMA's heavy reliance on learned value functions for advantage estimation contributes to its  performance, due to the gradient bias introduced as a result of approximation errors in the Q function \cite{gae}.  Finally, we found that COMA-indiv, which is essentially a naive implementation of COMA to an individual reward scenario, was highly unstable and failed to make progress towards solving the task, despite extensive hyper-parameter search.  We hypothesize that this issue is again due to biases introduced during advantage estimation, which compounded due to having one Q function estimate for every individual agent, instead of just one for the entire group.  

These results, and in particular the results of COMA and COMA-indiv, highlight two important advantages that PRD-AC offers over COMA.  The first is flexibility in advantage estimation strategy; whereas COMA is tethered to a particular (biased) advantage estimation strategy (which may fail in certain situations, as indicated by COMA-indiv), PRD is agnostic to advantage estimation strategy.  PRD can therefore take advantage of recent advances in advantage function estimation, such as GAE \cite{gae} (which effectively balances bias and variance) and action-conditioned baselines \cite{action_dependent_baselines} (which offers further variance reduction while avoiding bias).  The second advantage of PRD is that it is able to make better use of individual rewards, when they are available, as indicated by the fact that COMA fails when we attempt to incorporate individual rewards.  PRD, on the other hand, is able to use this additional information to significantly improve learning efficiency.

\section{Conclusions}
\label{sec:conclusion}

In this paper, we address the credit assignment problem in MARL by studying how large cooperative MARL problems can be decomposed into smaller decoupled subproblems, which may be solved independently to yield a fully cooperative group-level solution.  Additionally, we demonstrated how this decomposition, which we refer to as partial reward decoupling (PRD), can be incorporated into an actor-critic algorithm to yield an efficient model-free MARL algorithm that we term PRD actor-critic (PRD-AC). We show that PRD-AC consistently outperforms actor-critic algorithms that naively attempt to maximize either individual reward (greedy-AC) or total group reward (shared-AC), on a diverse set of tasks that require varying degrees of cooperation.  Finally, we relate PRD to an existing state-of-the-art multi-agent actor-critic algorithm (COMA), and show that for problems in which individual rewards are available, PRD outperforms COMA by making better use of this additional information, and by allowing more advanced advantage estimation strategies.



\bibliographystyle{IEEEtran}
\bibliography{refs}
\end{document}